\documentclass[11pt]{article}
\usepackage[final]{acl}
\usepackage[utf8]{inputenc}
\usepackage{amsfonts}
\usepackage{xcolor}
\usepackage[most]{tcolorbox}
\usepackage{times}
\usepackage{latexsym}
\usepackage{fvextra}
\usepackage{amssymb}
\usepackage{amsmath}
\usepackage[T1]{fontenc}
\usepackage[utf8]{inputenc}
\usepackage{microtype}
\usepackage{inconsolata}
\usepackage{algorithm}
\usepackage{algpseudocode}
\usepackage{float}
\usepackage{graphicx}
\usepackage{booktabs}
\usepackage{multirow}
\tcbuselibrary{listings,breakable}
\usepackage{listings}
\usepackage{silence}
\DefineVerbatimEnvironment{PromptListing}{Verbatim}{
  breaklines=true,
  breakanywhere=true,
  fontsize=\footnotesize
}

\newtcolorbox{promptbox}[1]{
  breakable,
  colback=gray!3,
  colframe=gray!40,
  boxrule=0.4pt,
  arc=2pt,
  left=6pt,
  right=6pt,
  top=6pt,
  bottom=6pt,
  title={#1},
  fonttitle=\bfseries,
}

\lstdefinestyle{prompt}{
  basicstyle=\ttfamily\scriptsize,
  breaklines=true,
  columns=fullflexible,
  frame=single,
  rulecolor=\color{black},
  frameround=ffff,
  showstringspaces=false,
  keepspaces=true,
  tabsize=2,
  xleftmargin=0.3em,
  xrightmargin=0.3em,
  aboveskip=0.4em,
  belowskip=0.4em,
  lineskip=-0.5pt
}

\title{BIRDTurk: Adaptation of the BIRD Text-to-SQL Dataset to Turkish}

\author{
  Burak Aktaş\textsuperscript{1}, Mehmet Can Baytekin\textsuperscript{1}, Süha Kağan Köse\textsuperscript{1}, Ömer İlbilgi\textsuperscript{1}, \\
  \textbf{Elif Özge Yılmaz}\textsuperscript{2}, \textbf{Çağrı Toraman}\textsuperscript{2}, \textbf{Bilge Kaan Görür}\textsuperscript{1} \\
  \textsuperscript{1}Roketsan Inc., Artificial Intelligence Technologies Unit, Turkey \\
  \textsuperscript{2}Middle East Technical University, Computer Engineering Department, Turkey \\
  \texttt{burak.aktas@roketsan.com.tr},
  \texttt{can.baytekin@roketsan.com.tr} \\
  \texttt{kagan.kose@roketsan.com.tr}, 
  \texttt{omer.ilbilgi@roketsan.com.tr} \\ 
  \texttt{yilmaz.ozge\_01@metu.edu.tr},
  \texttt{ctoraman@metu.edu.tr},
  \texttt{kaan.gorur@roketsan.com.tr} \\
}

\begin{document}
\maketitle

\begin{abstract}
Text-to-SQL systems have achieved strong performance on English benchmarks, yet their behavior in morphologically rich, low-resource languages remains largely unexplored. We introduce \emph{BIRDTurk}, the first Turkish adaptation of the BIRD benchmark, constructed through a controlled translation pipeline that adapts schema identifiers to Turkish while strictly preserving the logical structure and execution semantics of SQL queries and databases. Translation quality is validated on a sample size determined by the Central Limit Theorem to ensure 95\% confidence, achieving 98.15\% accuracy on human-evaluated samples. Using \emph{BIRDTurk}, we evaluate inference-based prompting, agentic multi-stage reasoning, and supervised fine-tuning. Our results reveal that Turkish introduces consistent performance degradation---driven by both structural linguistic divergence and underrepresentation in LLM pretraining---while agentic reasoning demonstrates stronger cross-lingual robustness. Supervised fine-tuning remains challenging for standard multilingual baselines but scales effectively with modern instruction-tuned models. \emph{BIRDTurk} provides a controlled testbed for cross-lingual Text-to-SQL evaluation under realistic database conditions. We release the training and development splits to support future research.\footnote{Links to our datasets and source code are available at: \url{https://github.com/metunlp/birdturk}}
\end{abstract}

\section{Introduction}

Natural language interfaces to databases aim to democratize data access by enabling non-expert users to query structured data using everyday language. This vision has driven substantial progress in the Text-to-SQL field, supported by large-scale benchmarks such as WikiSQL \citep{zhong2017seq2sql}, Spider \citep{yu-etal-2018-spider}, and more recently, BIRD \citep{li2023can} and Spider 2.0 \citep{lei2025spider2}.

Despite this progress, current benchmarks remain overwhelmingly English-centric \citep{min-etal-2019-pilot, tuan-nguyen-etal-2020-pilot}. This limitation is particularly important for morphologically rich and syntactically divergent languages like Turkish. Most state-of-the-art Text-to-SQL models implicitly rely on the close syntactic alignment between English and SQL, as both follow a Subject-Verb-Object (SVO) order. This shared structure enables a relatively linear correspondence between input tokens and SQL constructs \citep{qin2022survey}. In contrast, Turkish exhibits an agglutinative morphology with a Subject-Object-Verb (SOV) word order \citep{oflazer1994two, eryigit2008dependency, umutlu-etal-2025-evaluating}, which disrupts this direct alignment and complicates semantic parsing \citep{dou2023multispider}.

\begin{figure}[t]
    \centering
    \includegraphics[width=\linewidth]{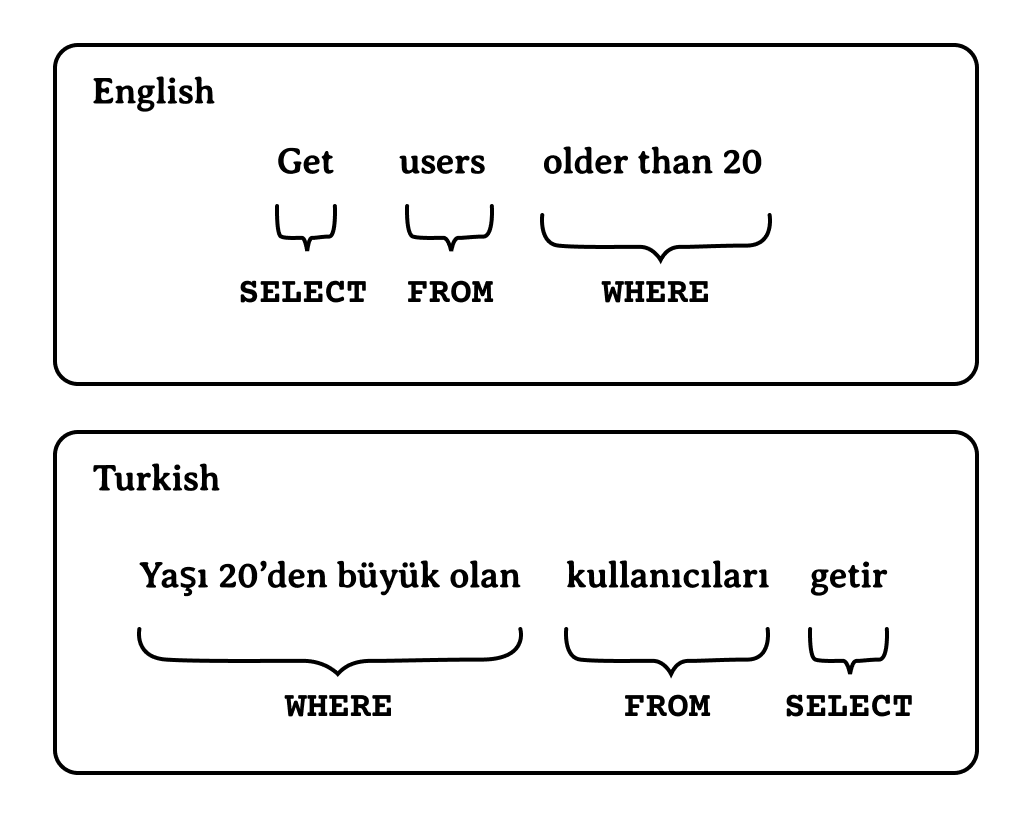}
    \caption{Structural divergence between English and Turkish queries. While English aligns linearly with SQL, Turkish distributes logic across suffixes and alters word order (SOV), complicating slot alignment.}
    \label{fig:comparison}
\end{figure}

For instance, consider the query ``\texttt{SELECT * FROM Users WHERE age > 20}''. As illustrated in Figure \ref{fig:comparison}, the English phrase ``\textit{Get users older than 20}'' aligns sequentially with the SQL logic. Conversely, in the Turkish translation ``\textit{Yaşı 20'den büyük kullanıcıları getir}'', the action corresponding to \texttt{SELECT} (``\textit{getir}'') appears at the very end. Furthermore, the logic for the \texttt{>} operator is morphologically distributed across the ablative suffix ``\textit{-den}'' and the adjective ``\textit{büyük}''. As shown in Figure~\ref{fig:comparison}, such structural mismatches significantly challenge intent recognition and slot alignment mechanisms in multilingual models \citep{dou2023multispider, kanburoglu2024turspider}.

While efforts such as Tur2SQL \citep{kanburoglu2023tur2sql} and TURSpider \citep{kanburoglu2024turspider} have provided valuable initial resources, they do not reflect the scale, schema complexity, or ``dirty data'' characteristics of modern enterprise environments. Existing Turkish datasets largely adhere to earlier benchmark paradigms, lacking the reasoning depth required to evaluate LLMs' capabilities in handling real-world database ambiguities as introduced by BIRD \citep{li2023can}.

To address this gap, we present \emph{BIRDTurk}, the first Turkish adaptation of the BIRD benchmark \citep{li2023can}. BIRDTurk is constructed via a controlled translation pipeline that preserves the logical structure and execution semantics of the original SQL queries and databases, while translating natural language questions and systematically localizing schema identifiers into Turkish. To ensure scalability and reliability, we employ a Central Limit Theorem (CLT)-based statistical verification framework, providing explicit confidence intervals for translation quality.

Our contributions can be summarized as follows:

\begin{itemize}
    \item We introduce \emph{BIRDTurk}, the first Turkish Text-to-SQL dataset adapted from BIRD.
    
    \item We propose a statistically grounded, CLT-based framework for validating large-scale dataset translations efficiently.
    
    \item We establish baseline results through systematic experiments spanning inference-based prompting, agentic reasoning, and supervised fine-tuning.
\end{itemize}

\section{Related Work}

\subsection{Evolution of Text-to-SQL Benchmarks}

The trajectory of Text-to-SQL benchmarks reflects a paradigm shift from constrained semantic parsing to real-world database grounding. Early datasets like ATIS and GeoQuery \citep{price-1990-evaluation, zelle1996learning} were limited to single domains, leading to overfitting and memorization issues \citep{finegan-dollak-etal-2018-improving}. While WikiSQL \citep{zhong2017seq2sql} introduced scale, it oversimplified the task to single-table operations. Spider \citep{yu-etal-2018-spider} addressed these limitations by introducing complex SQL structures (e.g., nesting, JOINs) and unseen schemas, establishing the \textit{de facto} standard for cross-domain structural generalization. However, these benchmarks operate on ``clean'' schemas, primarily testing a model's ability to map natural language tokens to SQL syntax rather than reasoning over database content.

Recent benchmarks move beyond syntactic alignment toward execution-centric evaluation on ``dirty'' data. BIRD \citep{li2023can} represents this leap by introducing massive, noisy databases (33.4 GB) that require reasoning over \textit{external knowledge} (e.g., domain terminology, numeric calculation) rather than simple schema linking. This shift exposes a significant gap between human performance (92.96\%) and state-of-the-art LLMs, highlighting the necessity of content-grounded reasoning. Extending this trajectory, Spider 2.0 \citep{lei2025spider2} further redefines the task by adopting an agentic ``Code Agent'' paradigm, requiring models to debug queries and navigate enterprise workflows, thereby positioning Text-to-SQL as a multi-turn software engineering challenge rather than a single-turn translation task.

\subsection{Challenges in Low-Resource Languages}

The scarcity of non-English datasets has prompted numerous adaptations of Spider 1.0, ranging from question-only translations in Chinese \citep{min-etal-2019-pilot} to full schema localization in Vietnamese \citep{tuan-nguyen-etal-2020-pilot}, Russian \citep{bakshandaeva-etal-2022-pauq}, and Arabic \citep{almohaimeed2024ar}. While foundational, these benchmarks primarily test syntactic alignment on idealized schemas, often bypassing the ``dirty data'' challenges inherent in enterprise environments.

In the Turkish domain, Tur2SQL \citep{kanburoglu2023tur2sql} and TURSpider \citep{kanburoglu2024turspider} represent significant milestones. However, they highlight a critical bottleneck: English-centric LLMs frequently struggle with Turkish's agglutinative morphology, leading to schema hallucinations due to suffix-induced tokenization mismatches \citep{kanburoglu2024turspider}. Furthermore, by adhering to Spider's topology, existing Turkish datasets do not assess the content-grounded reasoning capabilities required for modern database applications, a gap BIRDTurk aims to fill. In addition, recent Turkish LLM benchmarking efforts such as TurkBench \citep{toraman2026turkbench} include evaluations of instruction-following capabilities; however, SQL-oriented reasoning is only marginally covered, and the benchmark does not target the structured, content-grounded reasoning over relational databases required in this task.

\section{Dataset Construction}

We construct \emph{BIRDTurk} by translating the publicly available training and development splits of \textsc{BIRD} into Turkish. Our goal is to keep the benchmark \emph{functionally identical} across languages: the underlying databases and SQL semantics must remain unchanged, while the natural-language interface is localized. Following prior cross-lingual Text-to-SQL benchmarks \citep{min-etal-2019-pilot, tuan-nguyen-etal-2020-pilot, bakshandaeva-etal-2022-pauq, almohaimeed2024ar}, we prioritize \emph{semantic equivalence}, \emph{schema fidelity}, and \emph{execution consistency}. Since LLM-based translation can drift---particularly for morphologically rich languages like Turkish \citep{kanburoglu2024turspider}---we use a schema-grounded pipeline that constrains edits that could break executability.

We adopt \emph{schema-only localization}: we translate database/table/column identifiers but do \emph{not} translate database cell values (e.g., quoted string literals in \texttt{WHERE} clauses). This avoids introducing a separate value-linking problem and simplifies execution-level comparisons.

\subsection{Schema Mapping}
Before translating questions, we establish a deterministic \emph{schema mapping} $\mathcal{M}$ for each database, defining a one-to-one correspondence between original English identifiers (tables/columns) and their Turkish counterparts. Fixing this vocabulary upfront constrains question translation, evidence localization, and SQL rewriting to a closed identifier set, preventing out-of-vocabulary terms.

We extract schema metadata (table names, column attributes, foreign keys) directly from the SQLite files and include database descriptions when available to resolve ambiguities. We translate identifiers with \texttt{gemini-2.5-flash}~\citep{comanici2025gemini25} under strict constraints: ASCII-only \texttt{snake\_case} identifiers and standardized recurring sub-terms (e.g., \texttt{movie\_popularity} $\to$ \texttt{film\_populerligi}, \texttt{first\_name} $\to$ \texttt{ilk\_isim}). The full schema-mapping prompt is provided in Appendix~\ref{app:prompt_schema_mapping}.

We also address rare instances of \textit{identifier collision}, where distinct English identifiers map to the same Turkish form. Although strictly limited to two columns in our dataset, we enforce \emph{database-local uniqueness} and resolve these collisions deterministically (e.g., via stable suffix rules), which also mitigates schema hallucination in non-English schema settings \citep{kanburoglu2024turspider, dou2023multispider}.

\subsection{Translation and Localization Pipeline}
\begin{figure}[t]
    \centering
    \includegraphics[width=\linewidth]{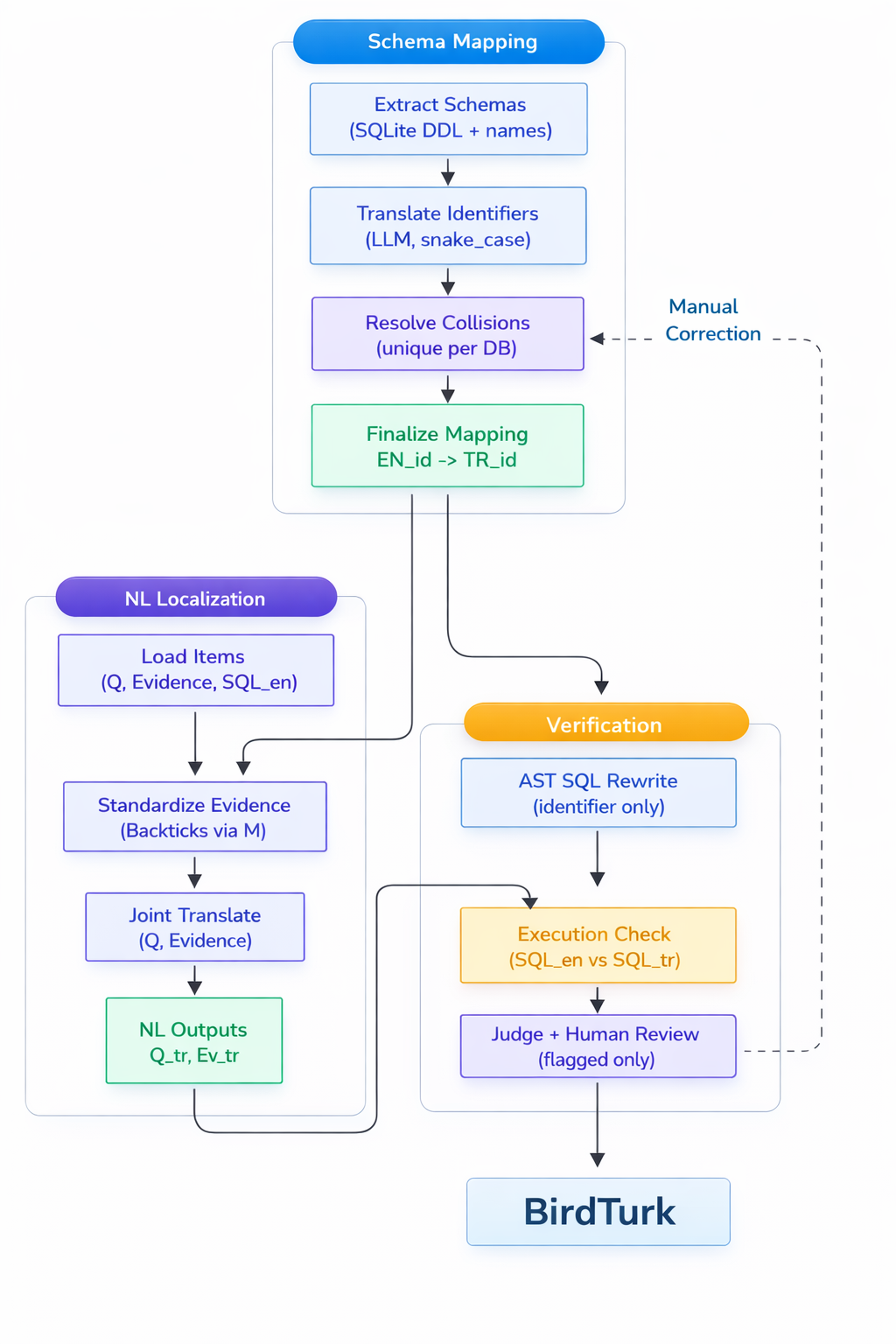}
    \caption{Translation and localization pipeline designed to convert the BIRD SQL benchmark into BIRDTurk while preserving SQL intent and enforcing Turkish language constraints.}
    \label{fig:pipeline}
\end{figure}

We frame localization as a constraint satisfaction problem in which the translated Turkish text must simultaneously preserve the original SQL semantics and comply with the constraints defined by $\mathcal{M}$. To achieve this, we design a three-stage translation and localization pipeline that systematically transforms the BIRD benchmark into its Turkish counterpart, BIRDTurk. The pipeline ensures semantic fidelity to the source queries while enforcing linguistic and structural consistency at each stage. The general workflow of this translation pipeline is illustrated in Figure~\ref{fig:pipeline}.

\subsubsection{Evidence Standardization and Schema Alignment}
Evidence fields in BIRD interleave natural language with schema references; inconsistencies can break the link between intent and executable SQL. Before translation, we deterministically rewrite all backticked identifiers in evidence using $\mathcal{M}$ and enforce an invariant: each backticked span must exactly match the Turkish \texttt{snake\_case} identifiers used in the localized SQL.

\subsubsection{Joint Question-Evidence Translation}
We then jointly translate the question and schema-aligned evidence in one context window using \texttt{gemini-2.5-flash} (prompt in Appendix~\ref{app:prompt_joint_translation}). The model must preserve backticked spans verbatim while localizing surrounding text, and it must retain semantic constants (numbers, dates) and comparative logic (e.g., \textit{highest/lowest/top-k}). For stylistic uniformity, we follow a fixed instruction set \citep{tuan-nguyen-etal-2020-pilot} (e.g., ``List'' $\to$ ``Listeleyiniz'', ``How many'' $\to$ ``Kaç \dots\ vardır?'').

\subsubsection{AST-Based SQL Localization}
To preserve execution behavior, we avoid neural SQL generation and apply a deterministic, structure-aware rewrite. We parse each SQL query into an Abstract Syntax Tree (AST)~\citep{aho2006compilers} and rewrite \emph{only identifier nodes} (tables/columns) via $\mathcal{M}$; all other elements (keywords, operators, functions, literals) remain unchanged. This avoids pitfalls of regex/string substitution (e.g., alias collisions, partial matches such as \texttt{table.column}, or accidental edits to literals) and preserves syntactic validity and semantic equivalence. As an additional safeguard, we use a rubric-based LLM-as-a-judge to verify text--SQL alignment for flagged instances (prompt in Appendix~\ref{app:prompt_llm_judge}).

\subsection{Quality Control and Statistical Validation}
To ensure reliability, we combine automated integrity checks with statistically grounded human evaluation.

\subsubsection{Automated Consistency Checks}
Before manual review, all examples undergo automated verification to filter structural and execution errors.

\paragraph{Execution Equivalence (Primary Signal).}
We adopt execution correctness as the primary validity criterion. For every instance, we execute both the original English SQL and the generated Turkish SQL on the underlying database. We verify that their result sets are identical ($R_{en} = R_{tr}$), strictly enforcing row ordering where specified. This step guarantees that the localization process preserves the executable semantics of the original query.

\paragraph{Structural Integrity and Schema Constraints.}
We algorithmically verify that all schema identifiers (tables, columns) referenced in the Turkish SQL strictly map to the localized schema. Furthermore, we cross-reference backticked identifiers in the natural language evidence against the localized schema to prevent hallucinated columns. Examples failing these checks are automatically flagged for correction.

\subsubsection{Statistical Semantic Validation}

Automated checks cannot fully capture fluency and semantic fidelity. To assess translation quality at scale, we use probabilistic sampling grounded in the Central Limit Theorem (CLT)~\citep{feller1968introduction}.

We model translation correctness as a Bernoulli random variable and estimate the required sample size for a \emph{95\% confidence level} ($Z = 1.96$) with an \emph{error margin of $\pm3\%$} ($E = 0.03$). Assuming maximum variance ($p = 0.5$), the sample size for an infinite population is:
\begin{equation}
    n_0 = \frac{Z^2 \cdot p(1-p)}{E^2} \approx \frac{1.96^2 \cdot 0.25}{0.0009} \approx 1{,}068
\end{equation}

Let $N$ denote the total number of samples in BIRDTurk, computed as the sum of the training and development splits ($N = 10{,}962$). Applying the Finite Population Correction (FPC) yields:
\begin{equation}
    n = \frac{n_0}{1 + \frac{n_0 - 1}{N}} \approx \frac{1{,}068}{1 + \frac{1{,}067}{10{,}962}} \approx 974
\end{equation}

Accordingly, we manually evaluated \emph{974 randomly sampled examples}; \emph{956 were judged correct}, yielding an observed translation accuracy of \emph{98.15\%}. Under the CLT, this implies the dataset-level accuracy lies within a $\pm3\%$ margin at 95\% confidence.

Algorithm~\ref{alg:translation-pipeline} summarizes our three-phase, schema-grounded construction pipeline, including schema mapping, joint question--evidence localization, and execution-consistent SQL rewriting with verification.

\begin{algorithm}[H]
\caption{Overview of the \textsc{BIRDTurk} construction pipeline.}
\label{alg:translation-pipeline}
\begin{algorithmic}[1]
\Require SQLite databases $\mathcal{D}$; English items $\mathcal{X}$ with $(db\_id, question, evidence, sql_{en})$
\Ensure Turkish items $\mathcal{X}_{tr}$ with $(db\_id_{tr}, question_{tr}, evidence_{tr}, sql_{tr})$

\State \textbf{Phase 1: Schema mapping}
\ForAll{$db \in \mathcal{D}$}
  \State $S_{db} \gets$ extract schema metadata (DDL + identifier list)
  \State $M_{db} \gets$ LLM translate identifiers in $S_{db}$ to Turkish ASCII \texttt{snake\_case}
  \State $M_{db} \gets$ resolve identifier collisions (unique within $db$)
\EndFor

\State \textbf{Phase 2: Natural language localization}
\ForAll{$x \in \mathcal{X}$}
  \State $M \gets M_{x.db\_id}$
  \State $e^{std} \gets$ rewrite backticked identifiers in $x.evidence$ via $M$
  \State $(q_{tr}, e_{tr}) \gets$ joint LLM translation of $(x.question, e^{std})$ with backticks frozen
\EndFor

\State \textbf{Phase 3: SQL localization and verification}
\ForAll{$x \in \mathcal{X}$}
  \State $M \gets M_{x.db\_id}$
  \State $sql_{tr} \gets$ AST rewrite $x.sql_{en}$ by replacing identifier nodes via $M$
  \If{\textsc{ExecEqual}($sql_{en}, sql_{tr}$ on $db$) is false}
    \State flag $x$ for review
  \EndIf
  \If{$x$ is flagged}
    \State rubric-based judge + human correction; update $M$ if needed
    \State re-run verification
  \EndIf
  \State store $(db\_id_{tr}, q_{tr}, e_{tr}, sql_{tr})$
\EndFor
\end{algorithmic}
\end{algorithm}

\subsection{Statistics of the Dataset}

BIRD is a large-scale benchmark with 12,751 text-to-SQL pairs across train/dev/test and 95 databases totaling 33.4~GB over 37 domains \citep{li2023can}. Each database contains 7.3 tables on average and roughly 549,000 rows; the largest database (``Donor'') is 4.5~GB.

\begin{table*}[t]
\centering
\caption{Linguistic and structural statistics for BIRDTurk Training and Development sets compared to the original English BIRD. The comparison highlights significant cross-lingual differences, such as reduced word counts due to agglutination and increased lexical diversity.}
\label{tab:combined_stats}
\resizebox{\textwidth}{!}{%
\begin{tabular}{lcccccc}
\toprule
\multirow{2}{*}{\textbf{Statistic}} & \multicolumn{3}{c}{\textbf{Training Set}} & \multicolumn{3}{c}{\textbf{Development Set}} \\ 
\cmidrule(lr){2-4} \cmidrule(lr){5-7}
 & \textbf{BIRD (En)} & \textbf{BIRDTurk (Tr)} & \textbf{Change (\%)} & \textbf{BIRD (En)} & \textbf{BIRDTurk (Tr)} & \textbf{Change (\%)} \\ 
\midrule
Total Questions & 9,428 & 9,428 & -- & 1,534 & 1,534 & -- \\
Avg. Words per Question & 14.05 & 10.21 & -27.3\% & 14.55 & 10.64 & -26.9\% \\
Avg. Characters per Question & 79.81 & 75.75 & -5.1\% & 82.76 & 79.18 & -4.3\% \\
Avg. Tokens per Question & 15.74 & 11.91 & -24.3\% & 16.24 & 12.22 & -24.8\% \\
\midrule
\multicolumn{7}{l}{\textit{Lexical Diversity}} \\
Vocabulary Size (Unique) & 9,002 & 15,142 & +68.2\% & 2,450 & 3,704 & +51.2\% \\
Type-Token Ratio (TTR) & 6.07\% & 13.49\% & +122.2\% & 9.84\% & 19.76\% & +100.8\% \\
\midrule
\multicolumn{7}{l}{\textit{Complexity \& Integrity}} \\
Avg. SQL Tokens & 31.03 & 30.89 & -0.5\% & 31.46 & 31.26 & -0.6\% \\
Avg. Evidence Tokens & 21.32 & 23.31 & +9.3\% & 21.49 & 22.04 & +2.6\% \\
\bottomrule
\end{tabular}%
}
\end{table*}

Table \ref{tab:combined_stats} presents a quantitative analysis of the linguistic characteristics of BIRDTurk and its structural alignment with the original BIRD benchmark. The comparison reveals systematic linguistic shifts that are consistent across both the training and development splits, confirming the robustness of the cross-lingual adaptation process.

Turkish agglutination compresses surface word counts (Train: \emph{-27.3\%}, Dev: \emph{-26.9\%}) as English particles are absorbed into suffixes, while character counts remain similar (4.3--5.1\% decrease), indicating preserved content at higher information density \citep{eryigit2008dependency}.

Lexical diversity increases sharply: TTR more than doubles (Train: \emph{+122.2\%}, Dev: \emph{+100.8\%}) and vocabulary size grows substantially (\emph{+68.2\%}, \emph{+51.2\%}). This reflects Turkish's productive morphology, which creates many surface forms from the same lemma and exacerbates sparsity for models \citep{hakkanitur2002statistical}.

SQL length remains effectively unchanged (-0.5\%), indicating preserved structural complexity. In contrast, evidence token counts increase (Train: \emph{+9.3\%}), consistent with the ``multilingual tokenization tax'' \citep{ahia2023do}: English-centric tokenizers often over-segment Turkish suffixes into multiple subword units \citep{rust2021how, toraman2022impact}.

\section{Experiments}

This section describes the experimental setup used to validate \emph{BIRDTurk} both as an evaluation dataset and as a supervised training resource. Following the experimental taxonomy of the original BIRD benchmark, we evaluate BIRDTurk under two complementary Text-to-SQL paradigms: inference-only prompting (including agentic reasoning pipelines) and supervised fine-tuning.

\subsection{Baseline Methods}

\subsubsection{Prompt-Based Inference}

We evaluate Text-to-SQL generation under inference-only settings, where no task-specific training or parameter updates are performed and all adaptation occurs through prompting.

As a direct baseline, we adopt a zero-shot in-context learning (ICL) setup, where the model receives a textual description of the database schema together with the natural language question and generates the SQL query in a single pass (see Appendix~\ref{app:prompt_bird_icl} and~\ref{app:prompt_birdturk_icl} for prompt templates). This setting assesses generalization to Turkish Text-to-SQL queries without explicit supervision or structural guidance.

To examine the effect of structured reasoning, we additionally evaluate DIN-SQL~\citep{pourreza2023din}, an agentic multi-stage inference pipeline that decomposes SQL generation into guided steps. The pipeline incorporates:
\begin{itemize}
    \item \textbf{Schema Linking} to ground relevant tables and columns,
    \item \textbf{Intent Classification} to infer high-level query structure,
    \item \textbf{Self-Correction} using execution feedback to refine generated SQL.
\end{itemize}
For Turkish compatibility, we translate the original DIN-SQL prompts using a structure-preserving strategy (Appendix~\ref{app:din_sql_translation_material}).

Both methods are instantiated using \texttt{gemini-2.5-flash-lite}, ensuring that performance differences arise from the reasoning strategy rather than model capacity. All inference experiments are conducted on both English BIRD and BIRDTurk under identical configurations.

\subsubsection{Supervised Fine-Tuning}

We also evaluate supervised Text-to-SQL learning on BIRDTurk to assess whether the translated dataset supports parameter-updated training.

Following the original BIRD setup, we fine-tune multilingual models from the mT5 family (\texttt{mT5-small}, \texttt{mT5-base}, \texttt{mT5-large}). While mT5 provides multilingual tokenization suitable for Turkish, fine-tuning these models yields limited performance gains, indicating challenges in learning effective Turkish Text-to-SQL mappings under this setup.

Motivated by these limitations, we transition to the \emph{Qwen2.5-Coder} family, utilizing the \texttt{0.5B}, \texttt{1.5B}, and \texttt{3B} instruction-tuned variants. Our selection of this specific architecture is driven by two complementary factors grounded in recent empirical findings. First, the Qwen2.5-Coder series achieves state-of-the-art performance in code generation benchmarks among open-weights models, benefiting from a massive pre-training corpus of 5.5 trillion tokens enriched with synthetic data~\citep{hui2024qwen2}. Second, and critical for our cross-lingual focus, recent research highlights the structural advantage of the Qwen architecture for Turkish. \citet{hacifazlioglu2024finetuning} demonstrate that the Qwen tokenizer achieves a superior compression ratio compared to models like Llama-3 and Gemma, effectively mitigating the over-segmentation of Turkish agglutinative suffixes. This tokenization efficiency allows the model to preserve the semantic integrity of Turkish natural language queries with fewer tokens, thereby enhancing the alignment between Turkish intent and SQL logic.

All Qwen models are fine-tuned and evaluated on BIRDTurk using the same supervised protocol. While their absolute performance remains below inference-based pipelines, they consistently outperform mT5 across all metrics, confirming that modern instruction-tuned models equipped with efficient tokenizers provide a more effective learning signal for Turkish Text-to-SQL tasks.

To ensure a fair comparison between base and fine-tuned capabilities, inference for both base and fine-tuned models is performed using the standard ICL prompting strategy.

\subsection{Evaluation Setup and Metrics}

We evaluate all models on the \emph{dev split} of the respective datasets, reporting metrics averaged over three independent runs to ensure robustness. The primary evaluation metrics are \emph{Execution Accuracy (EX)} and \emph{Valid Efficiency Score (VES)}, following the BIRD benchmark protocol, alongside \emph{Exact Match (EM)} for comparability with prior work.

\paragraph{Exact Match (EM).}
Exact Match measures whether the predicted SQL query is structurally identical to the ground-truth SQL after canonical normalization:
\[
\mathrm{EM} = \frac{1}{N} \sum_{n=1}^{N} \mathbb{I}_{\text{SQL}}(Y_n, \hat{Y}_n),
\]
where:
\[
\mathbb{I}_{\text{SQL}}(Y, \hat{Y}) =
\begin{cases}
1, & \text{if } Y \equiv \hat{Y} \text{ after normalization}, \\
0, & \text{otherwise}.
\end{cases}
\]

\paragraph{Execution Accuracy (EX).}
Execution Accuracy measures whether the execution results of the predicted SQL and the ground-truth SQL are identical:
\[
\mathrm{EX} = \frac{1}{N} \sum_{n=1}^{N} \mathbb{I}(V_n, \hat{V}_n),
\]
where:
\[
\mathbb{I}(V, \hat{V}) =
\begin{cases}
1, & \text{if } V = \hat{V}, \\
0, & \text{otherwise}.
\end{cases}
\]

\paragraph{Valid Efficiency Score (VES).}
Valid Efficiency Score extends Execution Accuracy by accounting for execution efficiency:
\[
\mathrm{VES} = \frac{1}{N} \sum_{n=1}^{N} 
\mathbb{I}(V_n, \hat{V}_n) \cdot R(Y_n, \hat{Y}_n),
\]
where:
\[
R(Y_n, \hat{Y}_n) = \sqrt{\frac{E(Y_n)}{E(\hat{Y}_n)}}.
\]
Here, $E(\cdot)$ denotes execution time. The square root term mitigates execution-time variance and extreme outliers~\citep{li2023can}.

\section{Results and Discussion}

\subsection{Overall Performance on BIRDTurk}

We first provide a high-level overview of model performance on the Turkish BIRDTurk dataset. 
Across all experimental settings, BIRDTurk constitutes a challenging testbed, reflecting both the linguistic properties of Turkish and the complexity of enterprise-scale Text-to-SQL tasks inherited from BIRD.

Three consistent trends emerge from our experiments.
First, supervised fine-tuning on BIRDTurk remains challenging for earlier multilingual baselines, while more recent instruction-tuned models exhibit clearer and more scalable learning behavior.
Second, inference-based approaches achieve substantially higher execution accuracy than supervised methods under identical evaluation conditions.
Third, within inference-based paradigms, agentic reasoning consistently improves over direct prompting across both English and Turkish.

We analyze these observations in detail below.

\subsection{Supervised Fine-Tuning Analysis on BIRDTurk}

We begin with supervised experiments to directly assess whether BIRDTurk supports parameter-updated learning in Turkish.
Table~\ref{tab:turkish_all} reports results for fine-tuned models evaluated on the same BIRDTurk development split.
Metrics are reported as Execution Accuracy (EA), Valid Efficiency Score (VES), and Exact Match (EM), where higher values indicate better performance ($\uparrow$).

\begin{table}[ht]
\centering
\setlength{\tabcolsep}{4pt}
\resizebox{\columnwidth}{!}{%
\begin{tabular}{l|ccc}
\hline
\multirow{2}{*}{\textbf{Model}} 
& \multicolumn{3}{c}{\textbf{Fine-Tuned} $\uparrow$} \\
\cline{2-4}
& \textbf{EA} & \textbf{VES} & \textbf{EM} \\
\hline

mT5-small 
& 0.26 & 0.28 & 0.00 \\

mT5-base  
& 0.98 & 1.04 & 0.06 \\

mT5-large 
& 1.83 & 2.05 & 0.13 \\

\hline

Qwen2.5-Coder-0.5B-Instruct 
& 1.24 & 1.38 & 0.33 \\

Qwen2.5-Coder-1.5B-Instruct 
& 7.24 & 8.01 & 2.61 \\

Qwen2.5-Coder-3B-Instruct 
& \textbf{15.38} & \textbf{17.12} & \textbf{4.82} \\

\hline
\end{tabular}
}
\caption{Supervised Text-to-SQL performance on the Turkish BIRDTurk dev split.
Results are reported for models after supervised fine-tuning.
Metrics include Execution Accuracy (EA), Valid Efficiency Score (VES), and Exact Match (EM), where higher is better ($\uparrow$).}
\label{tab:turkish_all}
\end{table}

Several discussions points to follow from Table~\ref{tab:turkish_all}.
We excluded base model results from the table as they all performed near 0\%, with the best performing base model (Qwen2.5-Coder-3B-Instruct) achieving only 2.22\% execution accuracy.
Multilingual baselines show only marginal improvements after fine-tuning, with execution accuracy remaining below 2\%.
In contrast, instruction-tuned models demonstrate clear and consistent gains, with performance scaling reliably with model size.

These results indicate that BIRDTurk provides a usable supervised signal; however, effective learning in Turkish benefits from models with stronger instruction-following and structured generation capabilities rather than multilingual coverage alone.

\subsection{Inference-Based Validation via English--Turkish Comparison}

To ensure that the observed supervised learning behavior is not an artifact of dataset construction or evaluation noise, we next examine inference-based performance under identical prompting conditions.
All inference experiments are conducted using the same underlying language model, \texttt{gemini-2.5-flash-lite}, enabling a controlled comparison across languages.

Table~\ref{tab:icl_compare} compares direct prompting (In-Context Learning) and agentic reasoning (DIN-SQL) on both the original English BIRD benchmark and its Turkish counterpart, BIRDTurk.

\begin{table}[ht]
\centering
\setlength{\tabcolsep}{4pt}
\resizebox{\columnwidth}{!}{%
\begin{tabular}{l|ccc|ccc}
\hline
\multirow{2}{*}{\textbf{Method}} 
& \multicolumn{3}{c|}{\textbf{BIRD (EN)} $\uparrow$}
& \multicolumn{3}{c}{\textbf{BIRDTurk (TR)} $\uparrow$} \\
\cline{2-7}
& \textbf{EA} & \textbf{VES} & \textbf{EM}
& \textbf{EA} & \textbf{VES} & \textbf{EM} \\
\hline

In-Context Learning
& 58.21 & 63.15 & 12.65 
& 43.16 & 45.34 & 10.82 \\

DIN-SQL 
& \textbf{60.89} & \textbf{64.66} & \textbf{17.60}
& \textbf{49.02} & \textbf{51.10} & \textbf{11.41} \\

\hline
\end{tabular}
}
\caption{Inference-based and agentic performance comparison on English BIRD and Turkish BIRDTurk dev splits.
Metrics include Execution Accuracy (EA), Valid Efficiency Score (VES), and Exact Match (EM), where higher values indicate better performance ($\uparrow$).}
\label{tab:icl_compare}
\end{table}

Across both datasets, English consistently outperforms Turkish in absolute terms.
However, the relative ordering of methods remains unchanged: DIN-SQL consistently improves over direct prompting in both languages.

This stability suggests that BIRDTurk preserves the core structural and reasoning characteristics of BIRD, and that the observed performance gap is largely attributable to language-induced difficulty rather than evaluation inconsistencies.

\subsection{The Impact of Agentic Reasoning Across Languages}

A closer examination of Table~\ref{tab:icl_compare} reveals an asymmetry in the relative gains provided by agentic reasoning across languages.
While DIN-SQL improves over direct prompting in both English and Turkish, the magnitude of improvement is more pronounced in the Turkish setting.

In English, agentic reasoning yields moderate but consistent gains across all evaluation metrics.
In contrast, for Turkish, DIN-SQL introduces larger relative improvements, particularly in Execution Accuracy and Valid Efficiency Score.
This pattern suggests that explicit task decomposition, schema grounding, and iterative correction play a more critical role when surface-level alignment between natural language and SQL is weaker.

One plausible explanation is that Turkish’s agglutinative morphology and SOV word order introduce additional ambiguity at the token and phrase level.
Agentic pipelines reduce reliance on direct token-to-SQL correspondence by enforcing intermediate reasoning steps aligned with schema structure and execution semantics.

Importantly, these trends are observed under identical inference configurations, indicating that the benefits of agentic reasoning stem primarily from linguistic factors rather than model-specific effects.

\subsection{Cross-Lingual Performance Comparison}

The remaining performance gap between English and Turkish can be attributed to two complementary factors: intrinsic linguistic properties of Turkish and representational imbalances in LLM pretraining.

From a linguistic perspective, Turkish poses structural challenges that complicate direct Text-to-SQL transfer: (1) a Subject--Object--Verb (SOV) word order that disrupts the linear alignment between natural language and SQL syntax; (2) agglutinative morphology leading to increased lexical sparsity, as semantic information is distributed across numerous low-frequency surface forms; and (3) suffix-heavy word forms that increase tokenization fragmentation, resulting in longer token sequences for equivalent semantic content.

Beyond these intrinsic challenges, current LLMs are predominantly trained on English-centric corpora, leaving Turkish significantly underrepresented in pretraining data. This imbalance limits the models' exposure to Turkish linguistic patterns, compounding the difficulties posed by the language's morphological complexity. The combination of these factors---structural divergence and limited pretraining coverage---creates a particularly challenging setting for cross-lingual generalization.

Agentic reasoning partially mitigates these effects by enforcing schema grounding and intermediate decision-making, thereby reducing reliance on surface-level token alignment and language-specific priors learned during pretraining.

\section{Conclusion}

In this work, we introduced \emph{BIRDTurk}, a Turkish adaptation of the BIRD Text-to-SQL benchmark designed for realistic database settings. By carefully controlling the adaptation process to maintain comparable execution behavior and database structure across languages, while localizing schema identifiers and natural language questions, we established a controlled evaluation dataset that enables direct and fair cross-lingual comparison.

Translation quality was validated using a statistically grounded CLT-based framework, providing dataset-level reliability without requiring exhaustive manual annotation. Leveraging BIRDTurk, we conducted a systematic evaluation of inference-based prompting, agentic multi-stage reasoning pipelines, and supervised fine-tuning approaches under Turkish linguistic conditions.

Our results demonstrate that Turkish introduces a consistent yet bounded performance degradation in inference-only settings. In contrast, agentic reasoning pipelines exhibit stronger robustness across languages, indicating that explicit task decomposition, schema grounding, and intermediate reasoning steps effectively mitigate language-induced challenges. While supervised fine-tuning remains difficult for standard multilingual baselines, more recent instruction-tuned models show clearer and more scalable learning behavior in the Turkish setting.

Beyond its current scope, BIRDTurk provides a foundation for several important research directions. Future extensions include the construction of a fully native Turkish Text-to-SQL benchmark to complement translation-based evaluation, the development of hybrid datasets combining translated and natively authored Turkish questions, and the expansion of BIRDTurk toward more agentic, enterprise-oriented, and multi-turn evaluation settings inspired by Spider~2.0. These directions would further strengthen the benchmark’s realism and broaden its applicability to practical deployment scenarios.

Overall, BIRDTurk fills a critical gap in Turkish Text-to-SQL research by enabling controlled cross-lingual evaluation and systematic analysis of modeling paradigms under realistic database conditions, while also laying the groundwork for more advanced and native Turkish evaluation frameworks.

\section*{Limitations}

Despite its contributions, BIRDTurk has several limitations that should be considered when interpreting the results.

\begin{itemize}
    \item BIRDTurk is constructed via translation rather than native Turkish question authoring. While this enables controlled cross-lingual comparison with BIRD, it may not fully reflect naturally occurring Turkish query formulations.
    \item Translations rely on a single LLM (Gemini), which may introduce systematic stylistic biases despite postprocessing and CLT-based validation.
    \item CLT-based validation provides dataset-level quality guarantees but does not replace exhaustive manual annotation at the individual sample level.
    \item Certain Turkish-specific linguistic phenomena (e.g., ellipsis, pragmatic inference) are likely underrepresented due to the translation-based construction process.
\end{itemize}

While translation noise is statistically bounded, a small subset of linguistically complex or highly implicit queries may still be affected.

\section*{Ethical Considerations}

\paragraph{Licensing and Copyright.}
BIRDTurk is constructed as a derivative work of the original BIRD benchmark \citep{li2023can}. We strictly adhere to the updated usage terms of the source material, which is distributed under the \emph{Creative Commons Attribution-ShareAlike 4.0 International (CC BY-SA 4.0)} license. In compliance with the "ShareAlike" provision, BIRDTurk is released under the same license to ensure that all derivative improvements remain open and accessible to the research community. We explicitly acknowledge the intellectual property rights of the original dataset creators and contribute this adaptation to foster reproducible and collaborative advancements in cross-lingual semantic parsing.

\paragraph{Use of Generative AI.}
Generative AI was used solely to assist with language editing. All scientific contributions, data construction and analysis, and interpretations presented in this work are original and were conducted entirely by the authors.

\section*{Acknowledgments}

We gratefully acknowledge support from Roketsan Inc. and the Google Gemini Academic Reward Program, which helped enable the experiments and computing resources used in this study.

\bibliography{acl_latex}

\clearpage

\appendix
\onecolumn

\section{Prompt Engineering for BIRDTurk}
\label{sec:appendix}

To ensure full reproducibility, we provide the complete set of prompt templates employed throughout our pipeline. These prompts are organized according to their functional roles within the system.

Prompts used for \emph{data translation and schema localization} enforce strictly defined JSON-only output formats and explicitly prohibit any unintended or implicit modifications to SQL tokens or schema identifiers.

Prompts used during \emph{inference and SQL generation} explicitly specify task-level constraints, business-logic guardrails, and SQLite-specific execution rules, thereby ensuring that the generated queries are both syntactically valid and semantically correct.

Finally, prompts used for \emph{DIN-SQL prompt translation} are designed as controlled drop-in replacements for training and evaluation. These prompts preserve the original structure and constraints of the DIN-SQL framework, while translating only the natural-language instructions without altering the underlying semantics.

\subsection{Schema Mapping Prompt}
\label{app:prompt_schema_mapping}

\begin{promptbox}{Schema Mapping}
\begin{PromptListing}

SYSTEM:
You are a careful data annotation assistant. Follow the rules exactly.

USER:
You will be given a single database schema package extracted from an SQLite file.
Your task is to produce a Turkish mapping for database, table, and column identifiers.

Return ONLY valid JSON with exactly these keys:
{
  "db_id_tr": "<ASCII-only lowercase snake_case>",
  "translations": {
    "<EN_IDENTIFIER>": "<TR_IDENTIFIER>",
    ...
  }
}

INPUT SCHEMA PACKAGE (JSON):
<<SCHEMA_PACKAGE_JSON>>

RULES (strict):
1) Scope: Translate ONLY identifiers (db/table/column names). Do NOT translate data values.
2) Output format:
   - db_id_tr: ASCII-only lowercase snake_case.
   - translations: ASCII-only lowercase snake_case for every TR_IDENTIFIER.
   - Turkish characters must be converted: ç->c, ğ->g, ı->i, ö->o, ş->s, ü->u.
3) Keep identifiers concise:
   - Do NOT add extra explanatory words.
   - Translate only what is explicitly present in the English identifier.
4) Consistency:
   - Translate recurring sub-terms consistently across the entire mapping (e.g., name, date, count, rate).
   - Keep common abbreviations unchanged when appropriate: id, api, ip, url, uuid, json, xml, http, https, sql.
5) Unknown acronyms:
   - If an identifier is an unknown acronym (e.g., "frpm"), keep it unchanged (still lowercase snake_case).
6) Uniqueness preference (best effort):
   - Try to avoid collisions within the same database.
   - If a collision seems likely, prefer adding a minimal context token rather than long phrases.
7) Do NOT output any extra keys, comments, or markdown. JSON only.
\end{PromptListing}
\end{promptbox}

\subsection{Joint Question--Evidence Translation Prompt}
\label{app:prompt_joint_translation}

\begin{promptbox}{Question--Evidence Translation}
\begin{PromptListing}

SYSTEM:
You are a careful translation assistant for a Text-to-SQL dataset. Follow the rules exactly.

USER:
You will be given:
- question_en: an English question
- evidence_std: an evidence text whose schema identifiers have ALREADY been standardized
  so that every schema identifier appears inside backticks as Turkish ASCII-only snake_case.
- (optional) sql_en and/or sql_tr for context (DO NOT rewrite SQL)

Your task:
Translate question_en and evidence_std into formal, fluent Turkish,
WITHOUT modifying anything inside backticks.

Return ONLY valid JSON with exactly these keys:
{
  "question_tr": "...",
  "evidence_tr": "..."
}

INPUT (JSON):
{
  "question_en": "<<QUESTION_EN>>",
  "evidence_std": "<<EVIDENCE_STD>>",
  "sql_en": "<<SQL_EN_OPTIONAL>>"
}

RULES (strict):
1) Backticks are read-only:
   - Keep anything inside `...` EXACTLY unchanged (no edits, no spacing changes, no casing changes).
   - Do NOT add or remove backticks.
2) Preserve meaning-critical tokens:
   - Do NOT change numbers, numeric ranges, units, or date formats.
   - Do NOT change quoted string literals if they appear in the text (e.g., 'Directly funded').
3) Preserve logical intent:
   - Comparative/superlative intent must remain correct (highest/lowest, most/least, top-k, at least/at most).
   - If question implies top-k, keep that intent explicit in Turkish.
4) Evidence structure must be preserved:
   - If evidence contains equations/ratios/operators (=, /, >, <, >=, <=), keep the same structure.
   - Translate only the surrounding natural language.
5) Formal Turkish style guide:
   - Prefer formal instructions: "Listeleyiniz", "Belirtiniz", "Gosteriniz", "Hesaplayiniz".
   - "how many" -> "Kac ... vardir?"
   - "average" -> "Ortalama ... nedir?"
6) Do NOT translate SQL tokens if they appear outside backticks:
   If the evidence text includes SQL keywords, clauses, operators, or function names,
   keep them EXACTLY as-is (case and spacing preserved). This includes:

   - Core clauses/keywords:
     SELECT, DISTINCT, FROM, WHERE, JOIN, INNER JOIN, LEFT JOIN, RIGHT JOIN, FULL JOIN,
     CROSS JOIN, ON, USING, GROUP BY, HAVING, ORDER BY, LIMIT, OFFSET,
     UNION, UNION ALL, INTERSECT, EXCEPT,
     WITH, AS, IN, NOT IN, EXISTS, NOT EXISTS,
     BETWEEN, LIKE, GLOB, IS NULL, IS NOT NULL, NULL,
     AND, OR, NOT,
     CASE, WHEN, THEN, ELSE, END,
     ASC, DESC

   - Aggregations/functions:
     COUNT, SUM, AVG, MIN, MAX,
     CAST, COALESCE, NULLIF,
     SUBSTR, INSTR, LENGTH, LOWER, UPPER, TRIM,
     ROUND, ABS,
     STRFTIME, DATE, DATETIME

   - Operators/punctuation (do not alter):
     =, !=, <>, >, <, >=, <=, +, -, *, /, %, ||, (, ), ,, .
7) Do NOT add new constraints or interpretations.
8) Output must be strict JSON with only the two keys above. No extra text.
\end{PromptListing}
\end{promptbox}

\subsection{LLM-as-a-Judge Rubric Prompt}
\label{app:prompt_llm_judge}

\begin{promptbox}{LLM-as-a-Judge}
\begin{PromptListing}

SYSTEM:
You are a strict evaluator for a Turkish Text-to-SQL dataset. You must output valid JSON only.

USER:
You will be given a Turkish Text-to-SQL item:
- question_tr
- evidence_tr (may be empty)
- sql_tr (localized SQL)
Optionally you may also see the original sql_en for reference.

Your task:
Evaluate whether the Turkish text (question_tr + evidence_tr) is semantically aligned with sql_tr.
Use the rubric below and output ONLY JSON.

Return ONLY valid JSON with exactly this schema:
{
  "intent_match":         {"pass": true/false, "reason": "..."},
  "constraints_preserved":{"pass": true/false, "reason": "..."},
  "aggregation_match":    {"pass": true/false, "reason": "..."},
  "ordering_limit_match": {"pass": true/false, "reason": "..."},
  "evidence_consistency": {"pass": true/false, "reason": "..."},
  "literal_handling":     {"pass": true/false, "reason": "..."},
  "overall_pass": true/false,
  "severity": "low" | "medium" | "high",
  "suggested_fix": "..."
}

INPUT (JSON):
{
  "question_tr": "<<QUESTION_TR>>",
  "evidence_tr": "<<EVIDENCE_TR>>",
  "sql_tr": "<<SQL_TR>>",
  "sql_en": "<<SQL_EN_OPTIONAL>>"
}

RUBRIC DEFINITIONS:
1) intent_match:
   - Does question_tr ask for exactly what sql_tr returns?
2) constraints_preserved:
   - Are implied filters/conditions reflected in sql_tr (and vice versa)?
3) aggregation_match:
   - If the text implies COUNT/SUM/AVG/MIN/MAX or grouping, does sql_tr match?
4) ordering_limit_match:
   - If the text implies highest/lowest/top-k, does ORDER BY/LIMIT/OFFSET match?
5) evidence_consistency:
   - If evidence_tr is present, are its backticked identifiers and computation consistent with sql_tr?
6) literal_handling:
   - Are quoted literals and constants handled consistently (no value translation/alteration)?

RULES (strict):
1) Do NOT rewrite sql_tr. Only evaluate it.
2) Reasons must be short (1-2 sentences each).
3) overall_pass:
   - True only if all critical dimensions pass.
   - evidence_consistency may fail only if evidence_tr is empty.
4) severity:
   - high: likely semantic mismatch or execution-breaking issue
   - medium: partial mismatch/ambiguity
   - low: mostly correct; minor issues
5) suggested_fix:
   - If overall_pass is true: "" (empty string).
   - If overall_pass is false: minimal fix or flag for manual review.
6) Output must be strict JSON only. No extra keys, no markdown.
\end{PromptListing}
\end{promptbox}


\subsection{In-Context Learning for BIRD}
\label{app:prompt_bird_icl}

\begin{promptbox}{In-Context Learning (BIRD)}
\begin{PromptListing}
  system_prompt: |
    You are an expert text-to-SQL model for the BIRD benchmark (business-domain databases).
    Given a natural language question, database schema, and optional hints, generate a single, correct SQLite SQL query.

    CRITICAL RULES FOR BIRD:
    1. NEVER use pre-calculated percentage/rate columns (e.g., "Percent (%) Eligible Free").
       ALWAYS calculate rates/percentages from base columns (e.g., "Free Meal Count / Enrollment").
       Example: For "eligible free rate", calculate: CAST(`Free Meal Count` AS REAL) / `Enrollment`
    
    2. Use EXACT column names from schema - check column descriptions carefully.
       If question mentions "school type" but schema has "Educational Option Type", use "Educational Option Type".
       Do NOT use similar-sounding columns - verify exact names from schema.
    
    3. Use EXACT filter values from value descriptions.
       If value description says "Continuation School" (not just "Continuation"), use the exact value.
       Check value descriptions in schema for correct filter values.
    
    4. ALWAYS read and apply hints when provided - they contain critical business logic.

    Key Guidelines:
    1. Schema Understanding:
       - Use ONLY tables and columns from the provided schema
       - Pay close attention to column descriptions and value descriptions
       - Check foreign key relationships for correct JOINs
       - Column names may contain spaces, use double quotes: "Column Name"
       - VERIFY exact column names - do not use similar-sounding columns
    
    2. Business Logic:
       - Understand business semantics (KPIs, rates, percentages, ratios)
       - Handle NULL values appropriately (exclude or handle with COALESCE)
       - Use CAST for proper type conversions (especially for percentages)
       - Apply business rules mentioned in hints
       - CALCULATE rates/percentages from base columns, NEVER use pre-calculated columns
    
    3. Complex Calculations:
       - For percentages: CAST(numerator AS REAL) / denominator * 100
       - For rates/ratios: CAST(count_A AS REAL) / count_B
       - Filter out zero/null denominators before division
       - Use CASE WHEN for conditional logic
       - Example: "eligible free rate" = CAST(`Free Meal Count` AS REAL) / `Enrollment`
    
    4. Query Optimization:
       - Use appropriate JOINs (INNER JOIN by default, LEFT JOIN when needed)
       - Apply filters early in the query (WHERE before HAVING)
       - Use subqueries for complex aggregations
       - Consider using CTEs (WITH clause) for readability if needed
    
    5. SQLite Specifics:
       - Use CAST(column AS REAL) for accurate decimal division
       - String literals use single quotes: 'value'
       - Column names with spaces or special chars use double quotes: "Column Name"
       - Use LIMIT with OFFSET for pagination
    
    Return ONLY the final SQL query with NO explanation, markdown, or additional text.
\end{PromptListing}
\end{promptbox}

\subsection{In-Context Learning for BIRDTurk}
\label{app:prompt_birdturk_icl}

\begin{promptbox}{In-Context Learning (BIRDTurk)}
\begin{PromptListing}
  system_prompt: |
    Sen Türkçe text-to-SQL dönüşümü için uzman bir modelsin (BIRDTurk benchmark, Türk iş dünyası veritabanları).
    Türkçe doğal dil sorusu, veritabanı şeması ve opsiyonel ipuçları verildiğinde, tek bir doğru SQLite SQL sorgusu üret.

    ÖNEMLİ: Tablo ve sütun isimleri TÜRKÇE'dir (ç, ğ, ı, ö, ş, ü karakterleri).
    
    KRİTİK KURALLAR:
    1. ÖNCEDEN hesaplanmış yüzde/oran sütunlarını ASLA kullanma.
       HER ZAMAN temel sütunlardan hesaplama yap.
       Örnek: "uygun ücretsiz oran" için: CAST(`Ücretsiz Yemek Sayısı` AS REAL) / `Kayıt`
    
    2. Şemadan TAM sütun isimlerini kullan - sütun açıklamalarını dikkatlice kontrol et.
       Soru "okul türü" dese bile şemada "Eğitim Seçenek Türü" varsa, onu kullan.
    
    3. Değer açıklamalarından TAM filtre değerlerini kullan.
    
    4. İpuçları verildiğinde HER ZAMAN oku ve uygula - kritik iş mantığı içerirler.

    Ana Kurallar:
    1. Şema Anlama:
       - Sadece verilen şemadan tablo ve sütunları kullan
       - Sütun açıklamalarına ve değer açıklamalarına dikkat et
       - JOIN'ler için foreign key ilişkilerini kontrol et
       - Boşluk içeren sütun isimleri için backtick kullan: `Sütun İsmi`
    
    2. İş Mantığı:
       - İş semantiğini anla (KPI'lar, oranlar, yüzdeler, rasyolar)
       - NULL değerleri uygun şekilde ele al
       - Uygun tip dönüşümleri için CAST kullan (özellikle yüzdeler için)
       - Oranları/yüzdeleri temel sütunlardan HESAPLA, önceden hesaplanmış sütunları ASLA kullanma
    
    3. Karmaşık Hesaplamalar:
       - Yüzdeler için: CAST(pay AS REAL) / payda * 100
       - Oranlar için: CAST(sayı_A AS REAL) / sayı_B
       - Bölmeden önce sıfır/null paydaları filtrele
    
    4. SQLite Özellikleri:
       - Doğru ondalık bölme için CAST(sütun AS REAL)
       - Boşluklu sütun isimleri: `Sütun İsmi` (backtick)
    
    Sadece final SQL sorgusunu döndür. Açıklama veya ek metin EKLEME.
\end{PromptListing}
\end{promptbox}

\subsection{DIN-SQL Prompt Translation Material}
\label{app:din_sql_translation_material}

\begin{promptbox}{DIN-SQL Translation}
\begin{PromptListing}
You are a senior NLP researcher and text-to-SQL benchmark expert.

USER:
Your task is to translate the DIN-SQL (BIRD) prompt text provided below from English into Turkish as a drop-in replacement for LLM training and evaluation.

STRICT RULES (apply to the text below):
1) Preserve meaning and structure exactly: keep headings, section order, numbering, bullets, delimiters, and any markdown/LaTeX formatting unchanged.
2) Translate ONLY natural-language sentences/phrases into fluent, professional, technical Turkish with consistent terminology.
3) Do NOT modify technical tokens unless they are clearly end-user natural language:
   - Keep SQL keywords, SQL code blocks, schema/table/column names, placeholders (e.g., {db_schema}), and symbolic notation unchanged.
4) Do NOT simplify, paraphrase, reorder, or reinterpret any instruction.
5) Do NOT add new examples, hints, or extra text. Keep safety/guardrail intent explicit.
6) OUTPUT: Return ONLY the full Turkish translation of the text below. No commentary, no additional formatting.

CONTENT TO TRANSLATE:
System Role: Senior NLP Researcher & SQL Expert
Objective: Given a Database Schema S and a Natural Language Question Q, generate a syntactically correct and semantically accurate SQL query Y.

Step 1: Schema Linking (Knowledge Extraction)
Identify the set of required tables T subset of S and columns C subset of S that satisfy the predicates in Q.
Map natural language entities to schema identifiers using external evidence E.

Step 2: Query Decomposition & Classification
Classify Q into complexity classes to determine the reasoning depth:
- Category I (Easy): Single table, basic filtering.
- Category II (Non-Nested): Joins, aggregations, ordering.
- Category III (Nested): Sub-queries, set operations, correlated joins.

Step 3: Logical Constraints and Guardrails
1) Schema Integrity: Use strictly provided identifiers in S.
2) Relational Validity: Ensure valid Foreign Key joins.
3) Aliasing Standard: Mandatory use of table aliases (e.g., T1, T2).
4) Output Exclusivity: Strictly contain the SQL code block.

Prompt Interface:
INPUT_SCHEMA: {db_schema}
INPUT_QUESTION: {question}
EXTERNAL_KNOWLEDGE: {evidence}

Expected Output Format:
```sql
SELECT T1.column FROM table AS T1 WHERE ...
\end{PromptListing}
\end{promptbox}

\twocolumn

\end{document}